\documentclass[sigconf,nonacm]{acmart}

\usepackage{caption}
\usepackage{subcaption}
\usepackage{algorithm}
\usepackage{algpseudocode}
\usepackage{multirow}
\usepackage{longtable}

\begin{document}
\title{Mining~Robust~Default~Configurations for~Resource-constrained~AutoML}

\author{Moe Kayali}
\email{kayali@cs.washington.edu}
\orcid{0000-0002-0643-6468}
\affiliation{
  \institution{University of Washington}
}
\authornote{Work performed as an intern at Microsoft Research.}

\author{Chi Wang}
\email{wang.chi@microsoft.com}
\affiliation{
\institution{Microsoft Research}
}


\begin{abstract}
 Automatic machine learning (Auto\textsc{ml}) is a key enabler of the mass deployment of the next generation of machine learning systems. A key desideratum for future \textsc{ml} systems is the automatic selection of models and hyperparameters. We present a novel method of selecting performant configurations for a given task by performing offline auto\textsc{ml} and mining over a diverse set of tasks. By mining the training tasks, we can select a compact portfolio of configurations that perform well over a wide variety of tasks, as well as learn a strategy to select portfolio configurations for yet-unseen tasks. The algorithm runs in a \textit{zero-shot} manner, that is without training any models online except the chosen one. In a compute- or time-constrained setting, this virtually instant selection is highly performant. Further, we show that our approach is effective for warm-starting existing auto\textsc{ml} platforms. In both settings, we demonstrate an improvement on the state-of-the-art by testing over $62$ classification and regression datasets. We also demonstrate the utility of recommending data-dependent default configurations that outperform widely used hand-crafted defaults.
\end{abstract}




\maketitle

\section{Introduction}

Machine learning (ML) is a key component of software infrastructure today. However, ML systems have been plagued with requirement of developer-hour-intensive tuning. They are very sensitive to the choice of preprocessing steps, models and hyperparameters, with small differences making the difference between a solution that meets its performance objectives and one that does not~\cite{DBLP:conf/aaai/LavessonD06}. Further still, the demand for ML applications has far outpaced---and is predicted to continue to outpace---the supply of experts capable of tuning such systems.

Automatic machine learning (AutoML) has been proposed as a solution to this problem. In the AutoML paradigm, the selection of the ML pipeline---the choice of preprocessing steps, hyperparameters and models---is itself approached as an optimization problem. This alleviates the need for the manual tuning of these components. If successful, AutoML will relieve a major burden on ML practitioners by enabling the automated selection of pipelines.

Typical autoML systems involve many trials of different configurations which consume a large amount of resources. 
More recently, the importance of the long tail of AutoML applications has been recognized~\cite{DBLP:conf/cidr/AgrawalCCFGIJKK20}. The number of developers utilizing ML is conservatively estimated at 10\% of the world in the next ten years which amounts to 20M engineering years.
They are predicted to build millions of ML-infused applications.
Two major differences mark these new, long-tail systems from the traditional mainstream ones. First, generalist developers build, deploy and maintain the models, rather than teams of machine learning experts. Second, it is not feasible to dedicate tremendous computational resources to these models. The resources available to many of these developers are personal laptops, public commons (\textit{e.g.}, Github Actions) and cloud compute credits. To make AutoML accessible to a majority of developers, addressing this setting is key.

The lowest resource bound for any autoML system is training at least one model. Can we build an autoML system component that recommends a good configuration for training a near optimal model for a giving learning task? We call such a system \textit{zero-shot}, in that no models are trained for the learning task at the recommendation time. A zero-shot autoML system has multiple benefits: low computational cost; low engineering complexity in continuous ML operations; easy to plug into existing ML workflow; reducing the risk of model selection bias due to overfitting the validation data or using incomplete training data; deployable in low resource environments or heterogeneous infrastructure; and scalable as microservices. The recommended configuration can also be used to warm start trial-based autoML systems when it is possible to run them.

Generally, no single configuration can perform well on all datasets. Thus the challenge of the zero-shot approach is selecting performant configurations for a dataset without knowing their performance.
The essential technique to overcoming this challenge is to perform offline autoML on a diverse set of training tasks and use the knowledge to cheaply decide a candidate configuration at runtime. While the offline autoML is still expensive, the experience learned from it is accumulated and leveraged to serve future online queries at nearly zero cost.

With this in mind, we present the following contributions:

\begin{enumerate}
    \item Propose a novel zero-shot autoML method, which by mining across a pool of training tasks, selects well-performing machine-learning models and hyperparameters on future tasks;
    \item Empirically validate our method, demonstrating an advance in the state of the art in the low-resource autoML setting, both as a zero-shot strategy and as a warm-starting strategy for an existent autoML framework;
    \item Demonstrate the utility in recommending data-dependent defaults which outperform hand-crafted defaults widely used;
    \item Examine the limitations of current approaches, providing evidence for the effectiveness of the remedies proposed in this work.
\end{enumerate}

\begin{figure}
    \centering
    \includegraphics[width=0.9\linewidth]{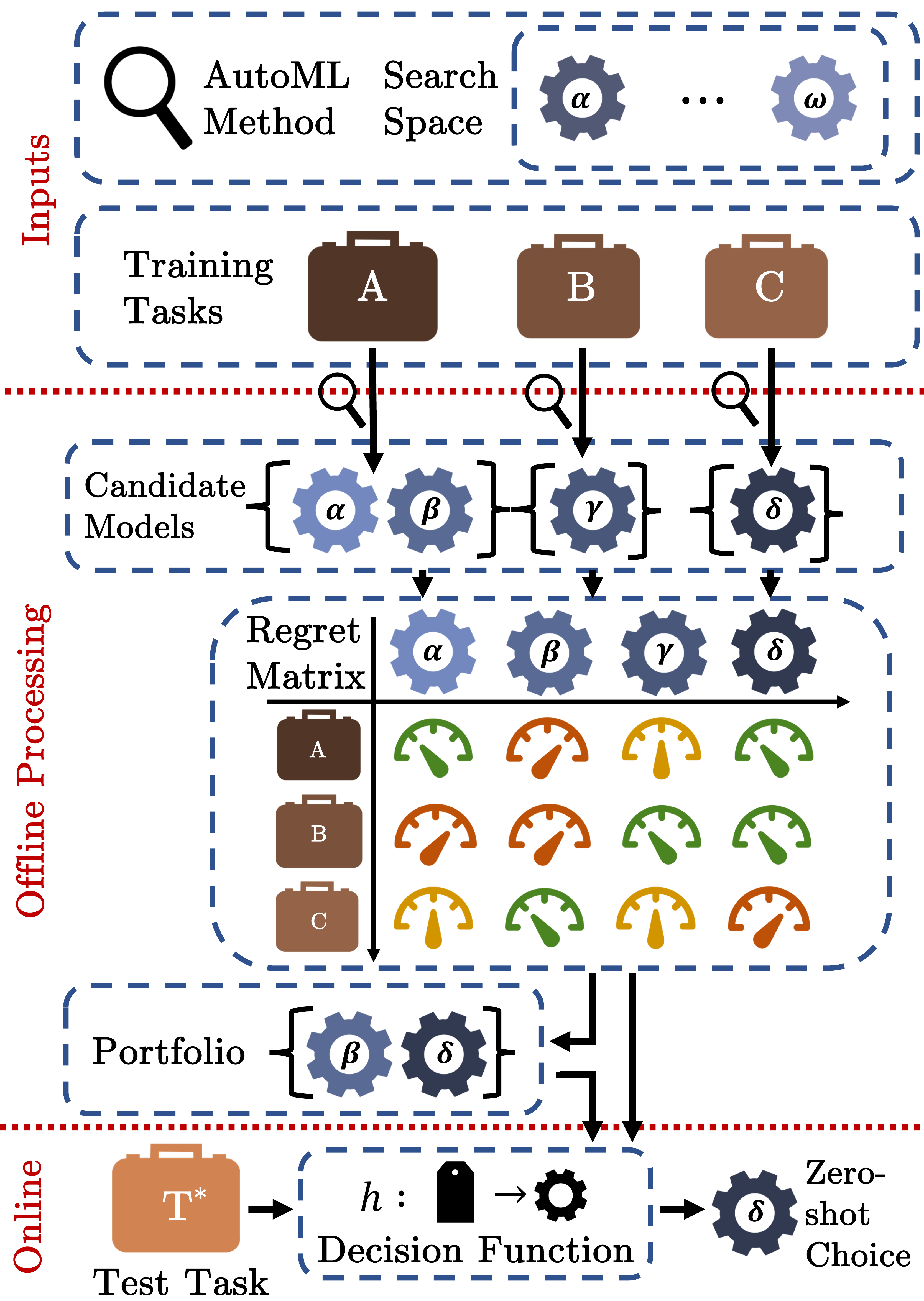}
    \caption{System architecture: given an auto\textsc{ml} method $A$, an infinite search space $S$ and training tasks $T$, we first mine candidate configurations $C$ out of each task. We then evaluate the performance of each $C, T$ combination to build a \textit{regret matrix}. This matrix is used to find a compact set of configurations, the \textit{portfolio}, from which we select a configuration at runtime using a decision function $h$.}
    \label{fig:architecture}
\end{figure}

\section{Background}
\label{section:background}

In our supervised setting, a \textit{task} is a $(X, y)$ pair on which we wish predict the label $y$ from the features $X$. The \textit{metafeatures} of task $t$ are statistical properties of $X$ and $y$, such as the number of features of $X$ or the distribution of classes in $y$. An \textit{automatic machine learning} (autoML) method \(A\) takes a task \(t\) as input and outputs $M$: preprocessing steps, machine learning models, their hyperparameters or any combination of these. In general, the output of $A$ must be specific to task $t$, as the no free lunch theorem~\cite{DBLP:journals/tec/DolpertM97} states that there is no $M$ that is well-performing on all $t$. In this work, we concern ourselves with autoML as model and hyperparameter selection. Our method can be easily applied to more complex pipeline selection if the output of the underlying autoML method is such complex pipelines.

An autoML approach considers a search space $S$ and a loss metric $L$ to minimize. In general, because $S$ is infinite, a criterion must be chosen to end the search. The most common criteria are: convergence, a time budget and the number of iterations. Given an automatic machine learning (autoML) method $A$ and a task \(t\), we use \(A(t)\) to denote the output of the autoML method after running for sufficient time. And we use \(L(A(t), t)\) to denote the loss of \(A(t)\) on \(t\). We consider the scenario where it takes unaffordable cost to run \(A\) until \(A(t)\) is obtained. We propose to learn from a collection of tasks $\{ t_i \}^N_{i=1} \in T$ a function of data-dependent default configuration \(h: T \rightarrow S\), which maps a task to a configuration in the search space. Once learned, the function \(h\) can be applied to a new task \(t\) and obtain a loss \(L(h(t), t)\) close to the loss of \(L(A(t), t)\), saving the expensive cost of running \(A\). Since the learning of such function is performed based on a collection of learning tasks and each learning task is represented by metafeatures, it can be characterized as a \emph{metalearning} technique. 

To measure performance in the resource-constrained setting in which we are interested, we introduce the notion of regret. 
We define the \textit{regret} as:
\begin{equation}
\label{eq:regret}
R(t) := L(h(t), t) - L(A(t), t)    
\end{equation}

We define the problem of \textit{robust default selection} as, given a regret bound $\varepsilon$, finding a minimal set of configurations $\{ C_j \}^m_{j=1}$ in $S$, called a \textit{portfolio}, and a \textit{decision function} $f : T \rightarrow [m]$ such that:
\[ h(t) = C_{f(t)}, R(t_i) \leq \varepsilon, \forall i \in N \]

We propose learning \(h\) on the minimal set rather than on $S$ as the latter is infinite and highly heterogeneous. Fitting a competitive $h$ over $S$ is difficult. While it is feasible in certain restricted settings~\cite{DBLP:conf/gecco/GijsbersPRBV21}, we are unaware of any successful attempts to improve the state-of-the-art with this approach. Instead, we opt to decomposing the mapping into two steps, each of which we are able to handle more easily.

We call an auto\textsc{ml} approach \textit{zero-shot} if evaluating $f$ does not involve training any models on the test task. If $f$ trains and evaluates only a few models, it can be referred to as \textit{$k$-shot}. Our problem formulation is novel, and we review related work in the next section.

\section{Related Work}
\label{sec:related}
Research in AutoML has landed in numerous open source or commercial software. Example of popular open source libraries include auto-sklearn~\cite{DBLP:conf/nips/FeurerKESBH15}, autogluon~\cite{agtabular}, FLAML~\cite{wang2021flaml}, H2O AutoML~\cite{ledell2020h2o}, and TPOT~\cite{Olson2016TPOT}. Examples of commercial end-to-end AutoML service include Amazon AWS SageMaker~\cite{DBLP:conf/sigmod/DasIBRGKDRPSWZS20}, DataRobot, Google Cloud AutoML Tables, Microsoft AzureML AutoML, Salesforce TransmogrifAI, H2O Driverless AI, Darwin AutoML and Oracle AutoML~\cite{DBLP:journals/pvldb/YakovlevMMCCVAK20}. We mainly review the research work which targets resource-constrained scenarios or uses metalearning.

Auto-Sklearn 1.0~\cite{DBLP:conf/nips/FeurerKESBH15} (\textsc{Askl} 1.0) is one of the first mainstream AutoML approaches, having won first place in the ChaLearn 2018 \cite{DBLP:conf/ijcnn/GuyonBCEEHMRSSV15} AutoML challenge.
This system uses metalearning, storing the best ML pipeline found by Bayesian optimization~\cite{hutter2011} in each task from a training set. 
At runtime, the metafeatures of the test task are matched to the \(k=25\) nearest neighbors in the training set. The pipeline with best accuracy on each nearest neighbor are tried on the test task. After that, Bayesian optimization is used to perform algorithm selection and hyperparameter optimization.

Auto-Sklearn 2.0~\cite{DBLP:journals/corr/abs-2007-04074} (\textsc{Askl} 2.0) is a more recent framework developed by the Auto-Sklearn 1.0 team. Despite the name, this system is a radically different approach rather than a refinement of its predecessor. Rather than relying on the nearest-neighbor selection algorithm, a fixed portfolio of pipelines is generated. Further, \textsc{Askl} 2.0 eschews the use of metafeatures completely. This is justified by the high cost of computing their choice of metafeatures, which resulted in timeouts for \textsc{Askl} 1.0 on the ChaLearn challenge. Instead, Auto-Sklearn 2.0 greedily builds a portfolio that maximizes average accuracy over all the training tasks assuming the best pipeline in the portfolio can be selected for each task. At runtime, the configurations in the portfolio are simply tried one after the other, then---if enough of the time budget is left---a Bayesian optimization proceeds.

Other works in this space include Oracle AutoML~\cite{DBLP:journals/pvldb/YakovlevMMCCVAK20}. In this system, the preprocessing, data reduction, model selection and hyperparameter tuning are done in one pass. While this cannot be characterized as a zero-shot system, as it repeatedly evaluates models online, the authors do pay special attention to the low time-budget case. At the same time, the algorithm relies on a highly parallel search at runtime, with experiments run on 36 cores. Metalearning is used in this system as a component to find a single hyperparameter configuration per learner. These configurations are used for ranking the learners per task for prioritization in hyperparameter tuning. The metalearning method is a simple variation of Auto-Sklearn 2.0, but for a different purpose.

Amazon SageMaker Autopilot~\cite{DBLP:conf/sigmod/DasIBRGKDRPSWZS20} is another AutoML system, with an emphasis on interactivity and interpretability, enabling the user to observe and influence the pipeline selection process. This approach utilizes metalearning and but assumes plentiful resources (as it is designed as a hosted service in Amazon's infrastructure) with a 10 hour time budget allocated for the system. Alpine Meadow~\cite{DBLP:conf/sigmod/ShangZBKECBUK19} is also an AutoML framework focused on speed and interpretability, returning initial results in well under a minute. However, similarly to SageMaker Autopilot, a highly-parallel (40 core, 80 thread) runtime environment is used as the testbed.

Other substantially different approaches~\cite{DBLP:conf/gecco/GijsbersPRBV21} include using metalearning to fit a symbolic function that maps the points in the metafeature space to hyperparameter values. While this approach is a zero-shot strategy, it does not allow for selecting a ML algorithm, only its hyperparameters. Further, it is unclear that fitting such a symbolic function for complex learners is viable, especially given that the hyperparameter search space is composed of heterogeneous subspaces~\cite{DBLP:journals/jmlr/BergstraB12}. According to the results in the paper, the symbolic defaults perform similarly to library defaults and nearest neighbor (as in \textsc{Askl}1) in real data.

We utilize the FLAML~\cite{wang2021flaml} framework as our primary testing framework. FLAML is a fast and lightweight auto\textsc{ml} library optimized for low computational resource consumption. It incorporates several off-the-shelf learners, picking between them using an adaptive search strategy. We choose FLAML due to its applicability to our resource-constrained setting.

\section{Approach}
\label{section:approach}

This section outlines our approach.
At a high level, we build a set of candidate configurations, evaluating their performance across a variety of tasks. We then pick a small set of configurations that cover many tasks well, picking between them at runtime using a nearest-neighbor query. This section describes the approach in detail; our strategy consists of three major components: configuration search, portfolio selection, and the runtime decision function. These components are illustrated in Figure~\ref{fig:architecture}.

\paragraph{Configuration search} This step finds good candidate configurations which form the basis of portfolio construction. The basic requirements for candidate configurations are: each candidate configuration needs to perform well on at least one task in $T$; and each task needs to have at least one candidate configuration that performs well on it. Otherwise, we will either have useless configurations and waste resources in evaluation or have insufficient coverage on the full set of tasks.
Fix an auto\textsc{ML} system $A$, which is formulated in Section~\ref{sec:related}. Run $A$ with a large time budget on each task in the training set $T_{\text{train}}$, until timeout or convergence. Denote $B = \{ A(t) \mid t \in T_{\text{train}} \}$. 
This set can serve as the candidate configurations.
When there are additional constraints, such as the training time per model, we run \(A\) again with that constraint on each task in the training set to get the candidate configurations.
Also, if there are external configurations that are deemed good candidate choices, such as the hand-crafted default configurations of each learner, they can be added to the candidate configuration set.
Denote the set of candidate configurations as $C$. Now, create a $|C| \times |T_{\text{train}}|$ performance matrix $P$ by evaluating the Cartesian product of model-hyperparameter configurations $C$ and tasks $T$. From this performance matrix, build a regret matrix $R$ by taking the difference between $P$ and $B$ column-wise, following Eq.~\eqref{eq:regret}. The computation of \(B\) and \(C\) are both embarrassingly parallelizable across different tasks \(T\), and the computation of \(P\) is embarrassingly parallelizable across \(C\times T\). With a good choice of \(A\) such that each autoML run is efficient using commodity hardware, this step can be scaled horizontally.

\paragraph{Portfolio Selection}

In this step, we prune the large number of configurations found in the search step into a smaller set, the \textit{portfolio}. Why not simply consider all configurations at runtime? We prefer the set be small because we posit, by Occam's razor, that such a set will generalize better on unseen tasks. For example, we could simply choose the set $B$ of \(N=|T_{train}|\) configs for the $N$ tasks, which corresponds to the \textsc{Askl} 1.0 strategy. This strategy would select a configuration that performs well on one dataset despite poor performance on all others. It is unlikely that such a configuration would generalize to unseen tasks. On the other extreme, we could select the one configuration with best mean regret. However, it is theoretically impossible for a single-configuration portfolio to perform well on all tasks~\cite{DBLP:journals/tec/DolpertM97}. In Section~\ref{section:results} we empirically evaluate both extremes, showing that the first approach does indeed overfit while no one configuration has acceptable worse-case regret. Instead, we want a few complimentary configurations, that cover the metafeature space well without overfitting to specific tasks.

We design our portfolio selection with these considerations in mind. This greedy algorithm is presented in Algorithm~\ref{algorithm:greedy}. We construct the portfolio in a bottom-up fashion, building it by starting with an empty set and adding one configuration at a time. An key feature of our algorithm is the error metric minimized: we minimize the \textit{sum-of-excess-regret} (\textsc{ser}). Given a target regret $\varepsilon$, we minimize the sum of the difference between the portfolio regret and $\varepsilon$ over all training tasks. When the target regret $\varepsilon$ has not been reached for any task, \textsc{Ser} behaves identically to a more traditional metric like mean regret. Once that threshold is reached for a given task, performance improvements on that task do not count towards \textsc{ser}.\footnote{However, if two portfolios have the same \textsc{ser} error, we prefer the one with lower mean regret.} The implication is that poor performance on a task cannot be compensated for by performing exceptionally well on another: by optimizing for the worst-case, we can increase the likelihood our porfolio covers the space of unseen tasks well.

Another detail of the algorithm is \textit{early stopping}. If we reach the target regret $\varepsilon$ or if adding a configuration to the portfolio does not decrease regret by at least a small numeric value, we terminate returning the unextended portfolio. The intuition behind the early stopping is to prevent overfitting in the attempt to reducing the regret to zero.

\begin{algorithm}
\caption{Greedy portfolio building with early-stopping}\label{alg:cap}
\label{algorithm:greedy}
\begin{algorithmic}
\Require $T, C$ \Comment{Training tasks, candidate configurations}
\Require $R : T \times C \rightarrow \mathbb{R}$ \Comment{Regret matrix}
\Require $\varepsilon : \mathbb{R}$ \Comment{Target regret}
\State $S \gets \varnothing$
\State $e \gets \infty$
\While{$C \neq \varnothing \wedge e > \varepsilon$}
\State $\mathbb{S} \gets \{S \cup \{c\} \mid c \in C\}$
\State $L \gets \{\sum_{t \in T} \max(\min_{s \in S} R(t, s) - \varepsilon, 0) \mid S' \in \mathbb{S} \} $ \Comment{\textsc{Ser} metric}
\If{$(1-\varepsilon / 2) e < \min L$} \Comment{Early stopping}
\State \textbf{break}
\EndIf
\If{$\min L$ is unique}
\State $S \gets \text{argmin}_{S' \in \mathbb{S}} L$  \Else \Comment{Break ties}
\State $L' \gets \{\sum_{t \in T} \min_{s \in S} R(t, s) / |T| \mid S' \in \mathbb{S} \} $
\State $S \gets \text{argmin}_{S' \in \mathbb{S}} L'$
\EndIf
\State $e \gets \min L$
\EndWhile
\State \Return $S$ \Comment{Portfolio}
\end{algorithmic}
\end{algorithm}

\paragraph{Decision Function}

At runtime, there remains the step of selecting which configuration in the portfolio to apply. To build a zero-shot system, we take a metafeature-driven approach. As such, the decision function can be considered a multiclass classifier, which takes as input the metafeatures of a task and selects one configuration from the portfolio. We experimentally validate (omitted here for brevity) three functions: nearest-neighbor, \textsc{svm}, and decision tree, and find nearest-neighbor the best performing. It has advantages in localized updates too.

\begin{figure}
    \centering
    \includegraphics[clip, trim=0.25cm 0.5cm 0.25cm 0.25cm, width=\columnwidth]{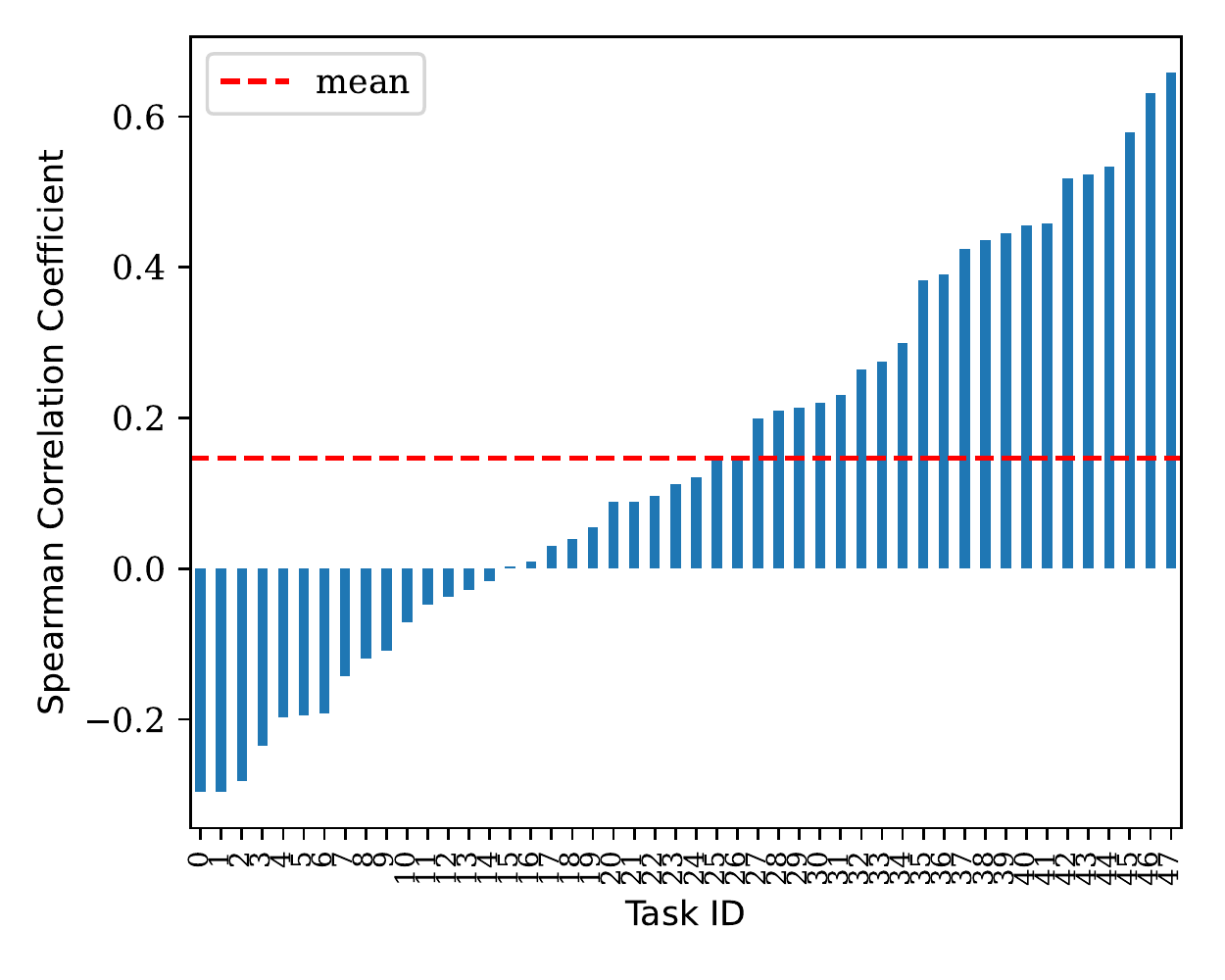}
    \caption{Ranking correlation between best configuration's transfer performance and similarity in metafeature space. For the majority of tasks, the ranking of the transfer performance for each other task's best configuration is positively correlated with the metafeature similarity. For a few tasks, the correlation is negative, implying room for improvement in terms of metafeature design.}
    \label{fig:metafeatures}
\end{figure}

As a design choice to study in this paper, we consider four simple metafeatures: (1) the number of instances, (2) the number of features, (3) the number of classes, and (4) the percentage of numeric features. The first two are well-known as important factors. All four metafeatures are selected to be fast to compute: in fact most of them are already computed by the data storage layer. This is in contrast to \textsc{Askl} 1.0's metafeatures, whose computation caused several timeouts at the ChaLearn challenge. We standardize the metafeatures before searching for the nearest neighbor.
These simple metafeatures work surprisingly well for the classification and regression tasks we studied. Figure~\ref{fig:metafeatures} plots preliminary evidence.
However, we stress that our overall framework is modularized and not fixated with this particular featurization choice. For example, when privileged information such as text descriptions are available for the tasks, one can use the technique in \cite{singh2021privileged} to compute text-based metafeatures.

\subsection{Discussion}

There are multiple practical advantages of our proposed approach at both online and offline stages.

At the online stage, the computation cost to find the zero-shot configuration is negligible, as a simple nearest-neighbor query. So it can be used to build a configuration recommendation microservice that scales easily.
The cost of obtaining a good model for a new task is 
just training one model, as no tuning is involved. That is the minimal cost one has to pay in autoML.
Without the need to setup a tuning loop infrastructure, the continuous integration and deployment of machine learning workflow becomes simpler. For different training data, the same existing user code for training can be reused, while different zero-shot configurations are automatically instantiated. And because no tuning is required, all training data are used for training and no validation data need to be generated for the tuning purpose. With that said, further tuning remains an option as the zero-shot configuration provides a good starting point.

The offline preparation can be customized for a domain and leverage the historical tuning data. For example, for a team building a database software, they can collect the training tasks and metafeatures from their application, and focus on learners and configurations that are suitable for their deployment requirement. With the three modularized components in our system, it is easy to perform such customization, e.g., by inserting filters of candidate configurations, changing metafeatures or even the decision function form. One important special case is the update of tasks/configurations. For example, a maintainer may encounter a new task where the zero-shot configuration is not good enough for deployment. They can fall back to performing a full tuning for the new task. At the same time, it is easy to add this new task and the tuned configuration to the input of offline preparation and upgrade the portfolio. Note that the update of candidate configurations and performance matrix in the first step can be performed incrementally and only the newly added task and configuration need to be evaluated against other configurations or tasks. The step of portfolio construction and decision function learning are both efficient after the update. So, the end-to-end update for the offline preparation is much less expensive than rerunning it from scratch. This enables an organization to share tuning experiences and improve the zero-shot solution over time collectively.

\section{Results}
\label{section:results}

\paragraph{Experimental setup} We experiment over $n=62$ tasks. We utilize a set of 35 classification tasks provided by the OpenML benchmark~\cite{amlb2019}. The benchmark contains a variety of classification tasks, both binary and multiclass, that are designed to represent a variety of workloads and so to be suitable for testing auto\textsc{ml} systems. However, as this benchmark only contains classification tasks, we extend it with 27 regression tasks curated in \cite{pmlr-v139-wu21d}. Of the $62$ tasks, $18$ are withheld as a validation set. This set, chosen through a randomized procedure, was blinded from the authors until algorithm design and all other experiments were concluded. We refer to the two sets of tasks as \textsc{cv} set (44 = 26 for classification and 18 for regression), and holdout set (18 = 9 for classification and 9 for regression).

We evaluate our approach by incorporating it into \textsc{Flaml}, the auto\textsc{ml} framework introduced in Section~\ref{section:background}. We use \textsc{Flaml} for candidate configuration search. Its search space consists of a choice of six learners \cite{DBLP:conf/kdd/ChenG16, DBLP:conf/nips/KeMFWCMYL17, DBLP:conf/nips/ProkhorenkovaGV18, scikit-learn}, for a total of 46 tunable hyperparameters\footnote{While the number of learners seems small, they include the modern and most competent libraries widely used by data scientists such as lightgbm and xgboost, and the search range of their hyperparameters is larger than other autoML libraries. So the search space is both complex enough and contains strong enough models.}. Each candidate configuration thus specifies one learner and the values of its hyperparameters. To standardize the tests in terms of instrumentation and preprocessing, we run all comparisons through the Open\textsc{ml} auto\textsc{ml} benchmarking framework~\cite{amlb2019}. Since we are motivated by the low resource setting, we use a much more frugal resource constraint (1 minute and 1 CPU core) than the original paper. The best known result with this low resource setting is reported in~\cite{wang2021flaml}, which we include as a baseline.

We compare against a baseline of the AutoSklearn 1.0 and 2.0 frameworks, described in Section~\ref{sec:related}. We note that both are complex, specialized systems---we adapt and evaluate the metalearning method proposed by them in our new problem setting. 
The original \textsc{Askl} 1.0 recommends the configuration obtained from the autoML run on the nearest neighbor task only.
We extend it to allow knowledge transfer between training tasks. 
This feature is present in \textsc{Askl} 2.0 and our algorithm. Without this change, \textsc{Askl} 1.0's performance would degrade, as the overfitting later described worsens.

\begin{table}
\caption{Regret on the 26 \emph{cv} classification tasks, each task 10-fold cross-validated. Lower is better.}
\label{table:classification}
\begin{tabular}{@{}lrrrrr@{}}
\toprule
  & \multicolumn{2}{c}{Zero-shot} & \multicolumn{3}{c}{Search or $k$-shot} \\ \cmidrule(rr){2-3} \cmidrule(r){4-6} 
           & Ours      & \textsc{Askl} 1.0     & \vtop{\hbox{\strut Ours+}\hbox{\strut Search}}     & Search    & \textsc{Askl} 2.0   \\ \midrule
Mean  & \textbf{0.0027}   & 0.0364   & \textbf{0.0021}      & 0.0134   & 0.0517   \\
Std. Dev.   & \textbf{0.0329}   & 0.0923   & 0.0411      & \textbf{0.0347}   & 0.0976   \\
Percentile \\
\quad 25\%  & \textbf{-0.0018}  & -0.0005  & -0.0035     & \textbf{-0.0001}  & 0.0024   \\
\quad 50\%  & \textbf{0.0019}   & 0.0047   & \textbf{0.0009}      & 0.0024   & 0.0128   \\
\quad 75\%  & \textbf{0.0102}   & 0.0276   &\textbf{0.0078}     & 0.0303   & 0.0556   \\
\quad 95\%  & \textbf{0.0344}   & 0.2785   & \textbf{0.0429}      & 0.0749   & 0.2975   \\
\quad 99\%  & \textbf{0.0854}   & 0.3454   & \textbf{0.0862}      & 0.1039   & 0.3740   \\
 \bottomrule
\end{tabular}
\end{table}

\begin{table}
\caption{Regret on the 18 \emph{cv} regression tasks, each task 10-fold cross validated. Lower is better.}
\label{table:regression}
\begin{tabular}{@{}lrrrrr@{}}
\toprule
  & \multicolumn{2}{c}{Zero-shot} & \multicolumn{3}{c}{Search or $k$-shot} \\ \cmidrule(rr){2-3} \cmidrule(r){4-6} 
           & Ours      & \textsc{Askl} 1.0     & \vtop{\hbox{\strut Ours+}\hbox{\strut Search}} & Search    & \textsc{Askl} 2.0   \\ \midrule
Mean & \textbf{0.0140}   & 0.0349  & \textbf{0.0025}  & 0.0082  & 0.0142  \\
Std. dev.  & \textbf{0.0239}   & 0.0716  & \textbf{0.0117}  & 0.0185  & 0.0384  \\
Percentile \\
\quad 25\% & 0.0004   & \textbf{-0.0007} & \textbf{-0.0002} & 0.0001  & 0.0000  \\
\quad 50\% & 0.0040   & \textbf{0.0017}  & \textbf{0.0009}  & 0.0014  & 0.0016  \\
\quad 75\% & \textbf{0.0201}   & 0.0394  & \textbf{0.0043}  & 0.0077  & 0.0086  \\
\quad 95\% & \textbf{0.0688}   & 0.1598  & \textbf{0.0168}  & 0.0462  & 0.1212  \\
\quad 99\% & \textbf{0.0777}   & 0.3640  & \textbf{0.0496}  & 0.0756  & 0.1774  \\ \bottomrule
\end{tabular}
\end{table}

\begin{table}
\caption{Regret on the 9 \textit{holdout} classification tasks, 10-fold cross validated. Lower is better.}
\label{table:classification-validation}
\begin{tabular}{@{}lrrrrr@{}}
\toprule
  & \multicolumn{2}{c}{Zero-shot} & \multicolumn{3}{c}{Search or $k$-shot} \\ \cmidrule(rr){2-3} \cmidrule(r){4-6} 
           & Ours      & \textsc{Askl} 1.0     & \vtop{\hbox{\strut Ours+}\hbox{\strut Search}}     & Search    & \textsc{Askl} 2.0   \\ \midrule
Mean & \textbf{0.0132}  & 0.0214  & \textbf{0.0024}      & 0.0238  & 0.0491 \\
Std. dev.  & \textbf{0.0299}  & 0.0465  & \textbf{0.0190}      & 0.0357  & 0.0468 \\
Percentile \\
\quad  25\% & 0.0003  & \textbf{-0.0017} & \textbf{-0.0067}     & 0.0011  & 0.0108 \\
\quad  50\% & 0.0086 & \textbf{0.0050}  & \textbf{0.0036}      & 0.0091  & 0.0347 \\
\quad  75\% & \textbf{0.0290}  & 0.0505  & \textbf{0.0084}      & 0.0291  & 0.0804 \\
\quad  95\% & \textbf{0.0722}  & 0.0785  & \textbf{0.0364}      & 0.1120  & 0.1355 \\
\quad  99\% & \textbf{0.0806}  & 0.1329  & \textbf{0.0463}      & 0.1303  & 0.1776 \\
 \bottomrule
\end{tabular}
\end{table}

\begin{table}
\caption{Regret on the 9 \textit{holdout} regression tasks, 10-fold cross validated. Lower is better.}
\label{table:regression-validation}
\begin{tabular}{@{}lrrrrr@{}}
\toprule
  & \multicolumn{2}{c}{Zero-shot} & \multicolumn{3}{c}{Search or $k$-shot} \\ \cmidrule(rr){2-3} \cmidrule(r){4-6} 
           & Ours      & \textsc{Askl} 1.0     & \vtop{\hbox{\strut Ours+}\hbox{\strut Search}}     & Search    & \textsc{Askl} 2.0   \\ \midrule
Mean & \textbf{0.0647}   & 0.0871  & \textbf{0.0462}  & 0.0513  & 0.0951  \\
Std. dev.  & \textbf{0.1151}   & 0.1509  & \textbf{0.1026}  & 0.1105  & 0.1821  \\
Percentile \\
\quad 25\% & \textbf{-0.0014}  & 0.0017  & \textbf{-0.0014} & -0.0001 & 0.0005  \\
\quad 50\% & \textbf{0.0001}  & 0.0068  & \textbf{0.0002}  & 0.0016  & 0.0074  \\
\quad 75\% & \textbf{0.0777}   & 0.1033  & 0.0215  & \textbf{0.0209}  & 0.0470  \\
\quad 95\% & \textbf{0.3031}   & 0.3784  & \textbf{0.2912}  & 0.3214  & 0.3838  \\
\quad 99\% & \textbf{0.3211}   & 0.5657  & \textbf{0.3456}  & 0.3597  & 0.8155 \\
 \bottomrule
\end{tabular}
\end{table}

\begin{table}
\caption{Percentage of failed folds out of 450 test folds.}
\label{table:timeout}
\begin{tabular}{@{}lrrrrr@{}}
\toprule
Strategy    &  Ours & Search & Ours+Search & Askl 1.0 & Askl 2.0\\ \midrule
Timed-out & 0.4\%   & 0.0\% & 2.4\%  & 8.6\% &  4.4\%          \\
\bottomrule
\end{tabular}
\end{table}

\subsection{Performance}

We compare the performance of our approach against several baselines in Tables~\ref{table:classification}--\ref{table:regression-validation}. We test two zero-shot approaches: ours and \textsc{Askl} 1.0. Further, we test two search or k-shot approaches: a \textsc{Flaml} search seeded with our zero-shot strategy, a standard \textsc{Flaml} search with hand-tuned defaults, and \textsc{Askl} 2.0. All zero-shot and search/$k$-shot approaches are allocated the same budget and access to the same metatraining data. Results are most comparable within the zero-shot group and within the search/$k$-shot group, as in the former the decision function does not have access to training at runtime (to retain the benefits we discussed earlier) while in the latter it does. The zero-shot recommendation consumes virtually none of the time budget, returning choices in the order of milliseconds.

We measure performance on regression tasks using the $R^2$ metric, while using the area under receiver operating characteristic curve (\textsc{roc-auc}) as the error metric for classification tasks. When classifying with more than two classes, we use a generalization of the \textsc{auc} metric to $n$ classes for computing the error metric~\cite{DBLP:journals/ml/HandT01}.

Performance on the \textsc{cv} set using leave-one-out cross-validation (\textsc{loo-cv}) is presented in Tables~\ref{table:classification}~and~\ref{table:regression}. In the \textsc{loo-cv} loop, each dataset in the \textsc{cv} set is used as a test task in turn. We metalearn on all tasks except the test task. We build a portfolio without performing any training or metalearning on the test task. Then we apply that portfolio on the test task and calculate the test regret. Within the \textsc{loo-cv} loop, we use a nested ten-fold cross-validation loop to calculate the error on each task. Finally, we report the mean, standard deviation, and percentiles on all the cross-validated folds (260 and 180 for classification and regression respectively).

Tables~\ref{table:classification-validation}~and~\ref{table:regression-validation} present performance on the holdout set of tasks. For this experiment, we build a portfolio using all datasets from \textsc{cv} set ($n=26, 18$ for classification and regression, respectively). We then evaluate the runtime performance in the resource-constrained setting using straightforward 10-fold cross validation. We emphasize that holdout datasets are never used in training or for metalearning.

In all tables, we only report dataset folds over which all five methods terminate. The sole reason for non-termination in all experiments is the selection of a model that cannot be trained in the benchmark, even after the test framework's attempts to reduce the runtime. Failures as a percentage of all the trials per auto\textsc{ml} strategy is presented in Table~\ref{table:timeout}. Both \textsc{Askl}~1.0~and~2.0 exhibit an order of magnitude larger number of failures compared to our zero-shot strategy. When using our strategy to seed the \textsc{Flaml} searcher, the number of time-outs increases, though still substantially less than the other baselines. We attribute this to increased number of steps where a timeout can occur: evaluating the zero-shot starting point, the random search, and the final retrain.

For completeness, we additionally evaluate the best-performing single-configuration portfolio. On the regression tasks, such a configuration has a mean regret of  $0.078$, a standard deviation of $0.256$, median regret of $0.014$,
75\% percentile regret $0.030$, 
95\% percentile regret of $0.173$, and finally a
99\% percentile regret of $0.964$. All are considerably worse than our method, as well as worse than most of the baselines. These results are expected in light of the theoretical background discussed in Section~\ref{section:background}.

\subsection{Ablation Experiments}

\begin{figure}
    \centering
    \includegraphics[width=0.75\linewidth]{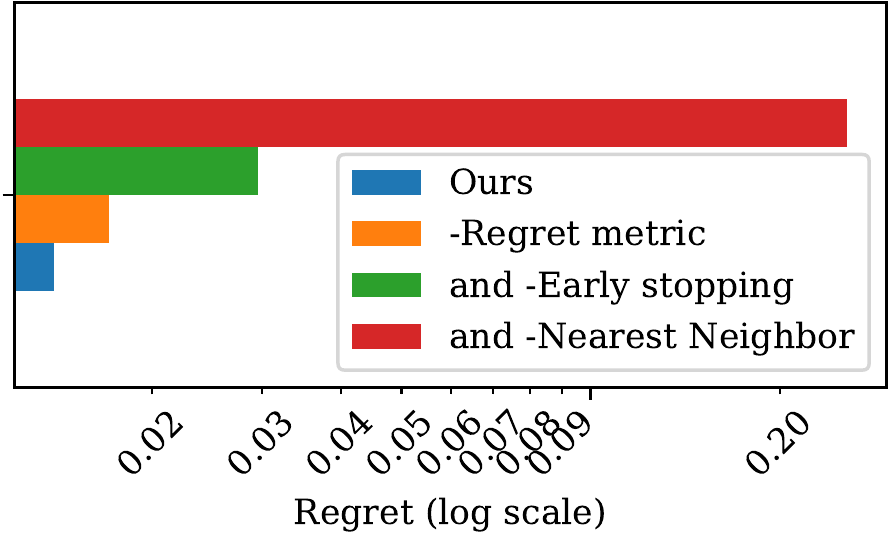}
    \caption{Ablation experiments, showing mean regret on $n=18$ test regression datasets. We measure the effect of disabling the regret metric, early stopping, and nearest neighbor search in the zero-shot setting.}
    \label{fig:ablation}
\end{figure}

Figure~\ref{fig:ablation} presents the results of the ablation experiments. We re-run our zero-shot algorithm, replacing our regret metric, disabling early stopping, and removing nearest neighbor as the runtime decision function. With each aspect of the algorithm omitted, we observe a increase in the mean test regret, supporting the inclusion of these aspects of our algorithm.

Replacing our sum-of-excess-regret metric with mean regret results in average increase in regret by 22\%. We note that the mean regret is also \textsc{Askl} 2.0's metric of choice. Further, disabling early stopping increases the likelihood of overfitting, as configurations will be added regardless of whether they decrease the error metric on training. This results in an \textsc{Askl} 1.0-like portfolio, adding 72\% to the regret of the previous configuration. Finally, as expected in the zero-shot setting, disabling nearest neighbor results in the most dramatic loss in performance. Unable at runtime to pick between models, we are forced to use a portfolio size of one, which increases regret by a factor of eight compared to the last ablation level.

\subsection{Scalability}

\begin{figure}
    \centering
    \includegraphics[width=0.75\linewidth]{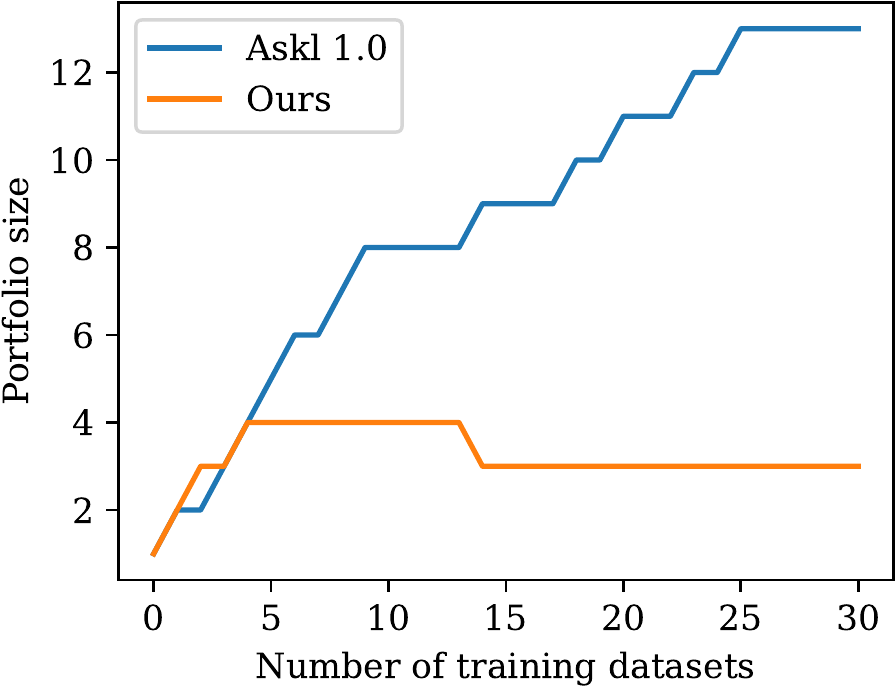}
    \caption{Portfolio size with increasing training set size. While the size of \textsc{Askl} 1.0's portfolio grows linearly as datasets are added in the metalearning step, our approach is able to find a ``core'' of configurations that perform well on the new tasks. Note that the smaller portfolio outperforms the larger one, as measured in Tables~\ref{table:classification},~\ref{table:regression}.}
    \label{fig:scalability}
\end{figure}

\begin{figure}
    \centering
    \begin{subfigure}{0.49\columnwidth}
    \centering
    \includegraphics[width=\linewidth]{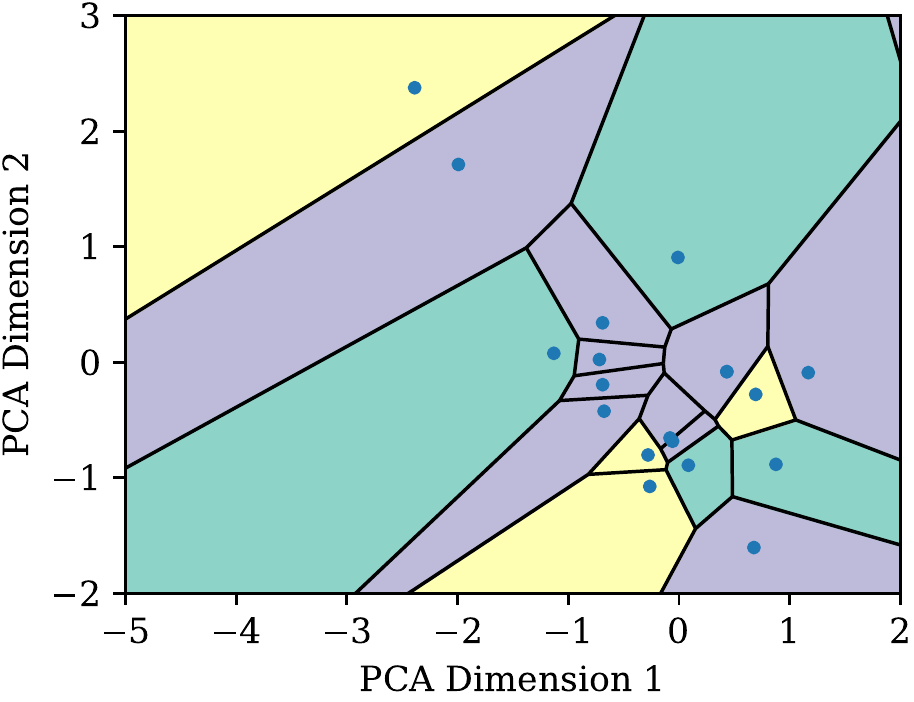}
    \caption{Our approach}
    \label{fig:voronoi_us_regression}
    \end{subfigure}
    \begin{subfigure}{0.49\columnwidth}
    \centering
    \includegraphics[width=\linewidth]{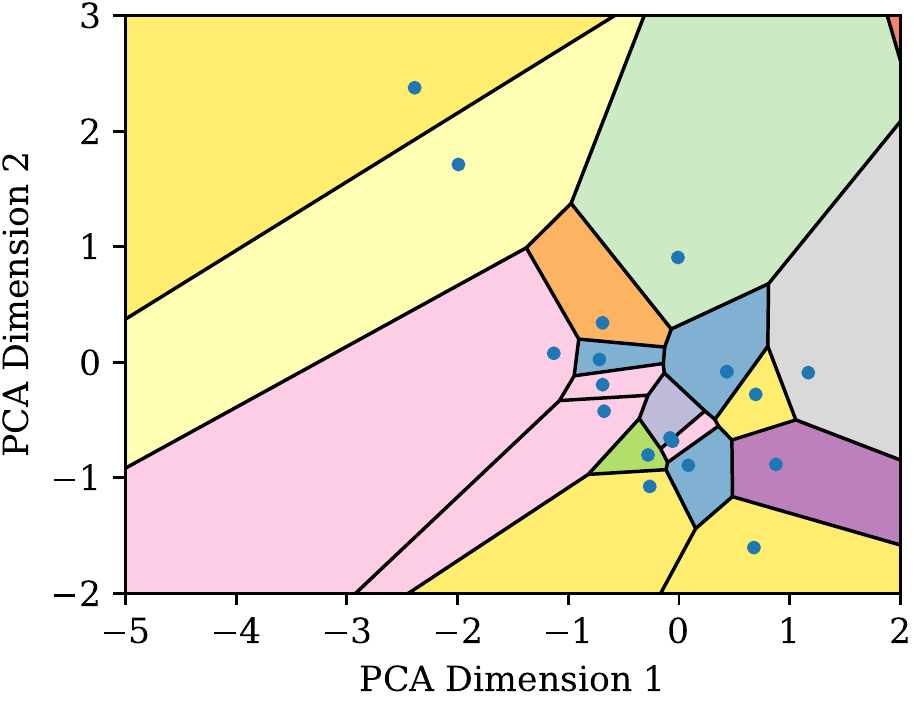}
    \caption{Autosklearn 1.0 
    }
    \label{fig:voronoi_others_regression}
    \end{subfigure}
    \caption{Comparison of the decision boundaries within the metafeature space of two metalearning strategies. Our portfolio~\ref{fig:voronoi_us_regression} outperforms the baseline portfolio~\ref{fig:voronoi_others_regression} despite being simpler (3 vs. 14 configurations). 
    }
    \label{fig:voronoi}
\end{figure}

We compare the number of configurations in each portfolio as number of training datasets is scaled up in Figure~\ref{fig:scalability}. \textsc{Askl} 1.0's approach results in a portfolio size linear in the number of training datasets. Meanwhile, our approach initially scales linearly for very small training set sizes---as each new configuration addresses a task the previous ones did not---but then quickly establishes a small set of configurations that cover the variety of training tasks. In fact, as the number of training sets grows past about ten, it is able to find more effective configurations and eventually \textit{shrink} the portfolio size. The final portfolio is over four times smaller than the \textsc{Askl} 1.0 portfolio yet, as seen in Tables~\ref{table:classification}-\ref{table:regression-validation}, outperforms it.

As noted in the beginning of Section~\ref{section:results}, we extended the original \textsc{Askl} 1.0 strategy with the ability to share knowledge about regret across training tasks. One effect is better portfolio scalability, as without this extension the portfolio size would scale in exact correspondence to the number of the training datasets, representing the line $x=y$ in Figure~\ref{fig:scalability}.

\subsection{Metafeature Space}

Figure~\ref{fig:voronoi} shows the decision boundaries of our approach versus \textsc{Askl}~1.0 on the regression tasks. In the Voronoi diagram, each point represents a training dataset. The decision boundaries are outlined in black, while the fill color indicates which portfolio configuration is used within a boundary. Two regions that map to the same configuration in the portfolio share the same color. For presentation purposes, the metafeature space was projected into two dimensions using principal component analysis (PCA).

At runtime, the metafeatures of the test task are used to map it into the polytope, deciding which configuration will be selected for that task. Each such area corresponds to one training dataset.

The metafeature space of \textsc{Askl} 1.0, in the Figure~\ref{fig:voronoi_others_regression} diagram, is comprised of 18 cells. These cells map to 14 unique configurations, meaning that only four regions share a configuration with another region. Figure~\ref{fig:voronoi_us_regression} shows the same diagram for our method. The metafeature space of our method differs: though the decision boundaries are the same (due to sharing the nearest-neighbor decision function) only three configurations are required to cover the 14 cells. Further, the cell coloring is generally smoother: most cells share a neighbor of the same color, unlike in \textsc{Askl} 1.0. Nonetheless, our simpler portfolio outperforms the more complex one. 

\begin{figure}
    \centering
    \includegraphics[width=0.85\linewidth]{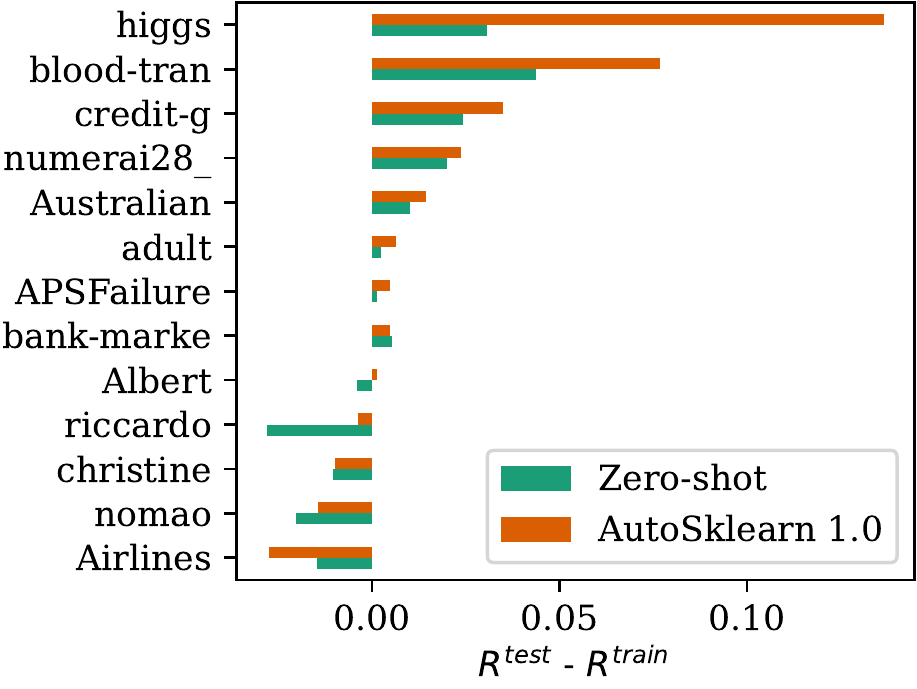}
    \caption{Overfitting by approach: difference between test and training regret for binary classification tasks. On average, AutoSklearn~1.0 underestimates the regret by a factor of ten compared to our zero-shot approach.}
    \label{fig:overfit-binary}
\end{figure}

\begin{figure*}
    \centering
    \includegraphics[width=\linewidth]{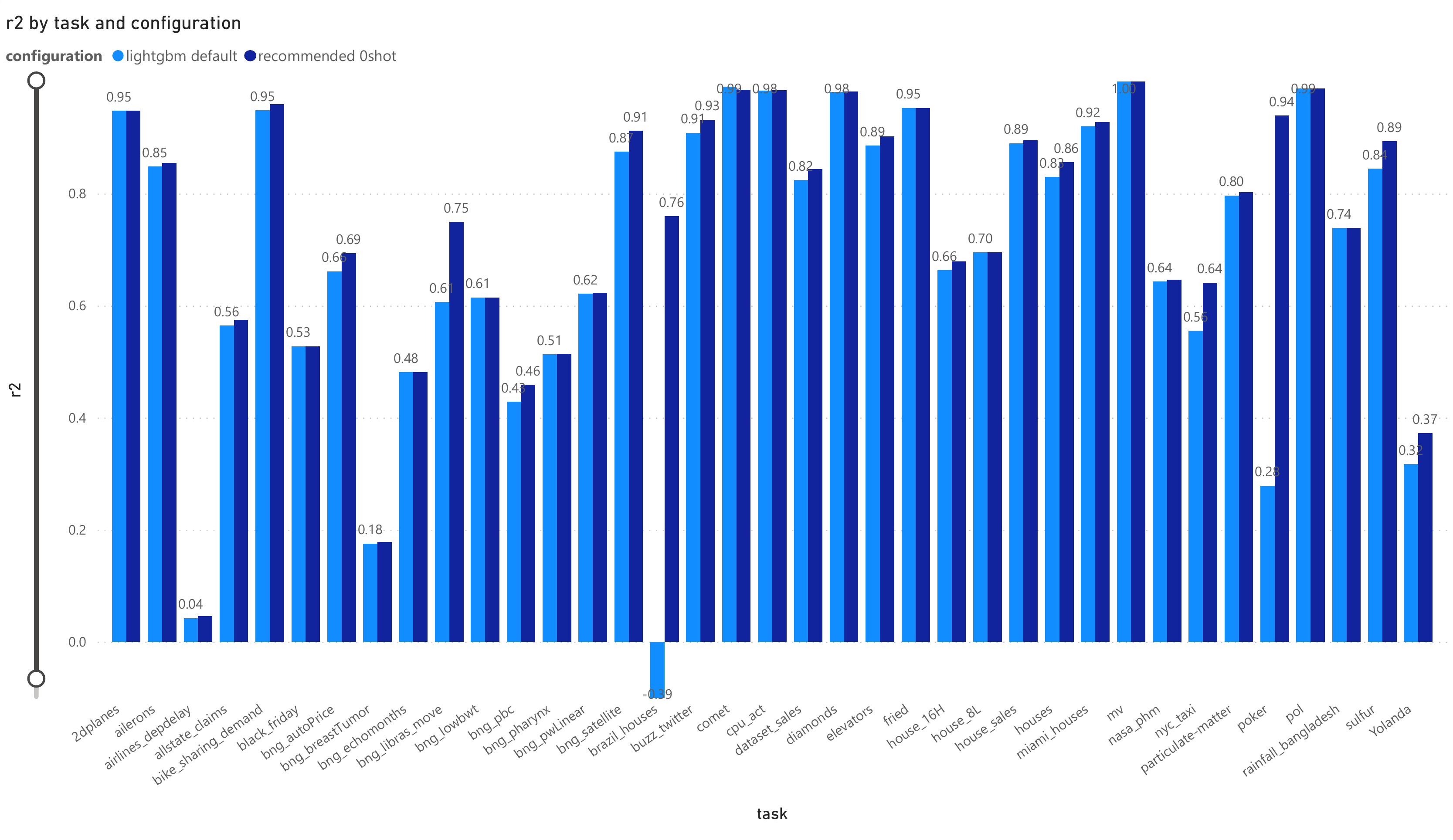}
    \caption{LightGBM performance with recommended zero-shot configurations and the library default.}
    \label{fig:lgbm}
\end{figure*}

\subsection{Portfolio Overfitting}
Figure~\ref{fig:overfit-binary} shows evidence of overfitting by \textsc{Askl} 1.0 at the portfolio build step. Compared to our approach, the difference between the mean regret at training time (the offline step) and observed regret on test (the online step) is approximately an order of magnitude larger. On average, \textsc{Askl} 1.0 \textit{under}estimates the regret by 0.019 versus our zero-shot approach's mean \textit{over}estimate of 0.002. On all datasets (except one, \texttt{riccardo}) our approach's estimate of the portfolio regret is closer. On \texttt{riccardo}, we underestimate the regret rather than overestimate it (too pessimistic). 

The overfitting for \textsc{Askl} 1.0 occurs because at each decision point in the algorithm, only one dataset is considered. Given the large number of datasets, it is highly likely that some of the ``good'' matches appear so by chance, rather than a genuinely well-performing and generalizable choices.
Our approach avoids falling in this trap by considering the performance of a configuration on all training tasks before selecting it for inclusion in the portfolio.
Its decision space is accordingly more regular in Figure~\ref{fig:voronoi_us_regression}.

While \textsc{Askl}~2.0 does not pick configurations that fail to generalize in the same manner as \textsc{Askl}~1.0, its portfolios lack diversity due to its error metric. Mean regret allows for a configuration that performs highly on an already well-covered task to be chosen, so long its performance on that task outweighs the weakness on other tasks. This results in \textsc{Askl}~2.0 having the largest regret variance of all tested approaches, as well as the largest tail regret, as seen particularly in Table~\ref{table:classification}. Our approach, by using the sum-of-excess-regret (\textsc{ser}) metric, is able to cover the worst case tasks well, resulting in a relatively lower variance and excellent tail performance. Additionally, the ablation experiments in Figure~\ref{fig:ablation} evidence that replacing \textsc{ser} by mean regret increases mean test regret by over 20\%. Examining a sample portfolio illustrates this: When APSFailure is the leave-one-out task, given the same set of configurations, our approach builds a portfolio of two LightGBM learners and one XGBoost learner while \textsc{Askl}~2.0 selects four LightGBM learners.

\section{Application}
\label{sec:application}
As an application of our technique, we study whether the recommended zero-shot configuration is better than the library default for popular and mature learners such as LightGBM, XGBoost and RandomForest. Each learner has offered a default configuration which is supposed to be robust across tasks as they have been adjusted by the library maintainers through many iterations of development cycles. These learners are widely used and testified by data scientists for many applications and evolved for many years. With such a high degree of maturity and a huge amount of crowd wisdom already built-in, chance is low for there to be a universally better default configuration. However, it is an intriguing question whether we can recommend data-dependent default configuration which is equal or better than the library default \textit{consistently}. It will not be surprising for this to be true in some tasks, but it is a tall order for an automatic zero-shot approach to outperform the carefully crafted default for every task. One recent study~\cite{DBLP:conf/gecco/GijsbersPRBV21} tried a symbolic approach and did not succeed in real data.

Figure~\ref{fig:lgbm} plots the 10-fold cross validated R\textsuperscript{2} score (the larger the better, 1 is perfect prediction, 0 is constant prediction with mean) for LightGBM on 37 regression tasks. 17 tasks are used for metalearning, and 20 are heldout: cpu\_act, bng\_autoPrice, elevators, house\_sales, bng\_satellite, allstate\_claims, black\_friday, buzz\_twitter, comet, rainfall\_bangladesh, bng\_libras\_move, dataset\_sales, diamonds, Yolanda, nyc\_taxi, bike\_sharing\_demand, sulfur, nasa\_phm, particulate-matter, miami\_houses. We observe that the recommended zero-shot configuration consistently outperforms or equalizes the library default configuration. The margin is large in quite a few tasks, such as bng\_libras\_move (+24\%), nyc\_taxi (+15\%) and Yolanda (+17\%). While the library default performs catastrophically in some tasks like brazil\_houses (-0.39 vs 0.76) and poker (0.28 vs. 0.94), the zero-shot approach stands strong in the worst case performance.
The only case where the recommended zero-shot is worse than the library default is on a heldout task comet, where the r2 scores are 0.9857 vs. 0.9907 with a -0.5\% margin. 
The mined portfolio contains three configurations obtained from autoML runs.
Overall, the portfolio demonstrates a robust performance against diverse tasks, and looks promising as a strong alternative to the static preset. We applied the same study to XGBoost, Randomforest and ExtraTrees and observed the same. We will share the mined portfolios and decision functions via \url{https://github.com/microsoft/FLAML}.

\section{Conclusion and Future Work}

This work presents a novel automatic machine learning approach, focused on a resource-constrained setting. By mining on a variety of tasks offline, we are able to assemble a small portfolio of configurations that perform well on a variety of datasets. Further, at runtime we can effectively choose a configuration from the portfolio in a \textit{zero-shot} manner; that is, without training any models except the chosen one. We evaluate this novel approach on $62$ datasets representing a variety of regression and classification tasks. It outperforms several baselines, including the state-of-the-art metalearning methods used in Auto\-Sklearn, on \textsc{cv} and holdout tasks. Performance is especially strong on tail tasks, which previous works neglect in this setting. Further, it is suitable for warm-starting auto\textsc{ml} frameworks, as well as recommending data-dependent default configurations to replace the fixed default settings from popular libraries. We identify short-comings of previous approaches---including portfolio overfitting, scalability problems and poor dataset coverage---and demonstrate the ability of our algorithm to address them.

In this work, we show that our approach is not restricted to one type of machine learning task: it is able to pick configurations for binary classification, multiclass classification and regression. A natural direction then is testing extensions to other task categories, for example time-series forecasting. It is also interesting to see the effect of a larger corpus of pipelines to apply metalearning.
Another possible improvement is an extension of the algorithm presented in Section~\ref{section:approach} to incorporate feedback from the decision function into the portfolio-building step. In the case of our decision function, nearest-neighbor, the greedy algorithm could take into account which configurations the nearest-neighbor matching would pick and adjust its portfolio accordingly. However, in preliminary experiments, we discovered that such feedback interferes with the early stopping condition. With this limitation addressed, the zero-shot performance could be improved further.

\begin{acks}
Special thanks to Anshuman Dutt for his extensive, helpful feedback throughout the research process. We are also grateful to DevDiv and ML.NET teams for their feedback on the key points of this work.
\end{acks}

\bibliographystyle{ACM-Reference-Format}
\bibliography{bibliography}


\begin{thebibliography}{24}


\ifx \showCODEN    \undefined \def \showCODEN     #1{\unskip}     \fi
\ifx \showDOI      \undefined \def \showDOI       #1{#1}\fi
\ifx \showISBNx    \undefined \def \showISBNx     #1{\unskip}     \fi
\ifx \showISBNxiii \undefined \def \showISBNxiii  #1{\unskip}     \fi
\ifx \showISSN     \undefined \def \showISSN      #1{\unskip}     \fi
\ifx \showLCCN     \undefined \def \showLCCN      #1{\unskip}     \fi
\ifx \shownote     \undefined \def \shownote      #1{#1}          \fi
\ifx \showarticletitle \undefined \def \showarticletitle #1{#1}   \fi
\ifx \showURL      \undefined \def \showURL       {\relax}        \fi
\providecommand\bibfield[2]{#2}
\providecommand\bibinfo[2]{#2}
\providecommand\natexlab[1]{#1}
\providecommand\showeprint[2][]{arXiv:#2}

\bibitem[\protect\citeauthoryear{Agrawal, Chatterjee, Curino, Floratou, Godwal,
  Interlandi, Jindal, Karanasos, Krishnan, Kroth, Leeka, Park, Patel, Poppe,
  Psallidas, Ramakrishnan, Roy, Saur, Sen, Weimer, Wright, and Zhu}{Agrawal
  et~al\mbox{.}}{2020}]%
        {DBLP:conf/cidr/AgrawalCCFGIJKK20}
\bibfield{author}{\bibinfo{person}{Ashvin Agrawal}, \bibinfo{person}{Rony
  Chatterjee}, \bibinfo{person}{Carlo Curino}, \bibinfo{person}{Avrilia
  Floratou}, \bibinfo{person}{Neha Godwal}, \bibinfo{person}{Matteo
  Interlandi}, \bibinfo{person}{Alekh Jindal}, \bibinfo{person}{Konstantinos
  Karanasos}, \bibinfo{person}{Subru Krishnan}, \bibinfo{person}{Brian Kroth},
  \bibinfo{person}{Jyoti Leeka}, \bibinfo{person}{Kwanghyun Park},
  \bibinfo{person}{Hiren Patel}, \bibinfo{person}{Olga Poppe},
  \bibinfo{person}{Fotis Psallidas}, \bibinfo{person}{Raghu Ramakrishnan},
  \bibinfo{person}{Abhishek Roy}, \bibinfo{person}{Karla Saur},
  \bibinfo{person}{Rathijit Sen}, \bibinfo{person}{Markus Weimer},
  \bibinfo{person}{Travis Wright}, {and} \bibinfo{person}{Yiwen Zhu}.}
  \bibinfo{year}{2020}\natexlab{}.
\newblock \showarticletitle{Cloudy with high chance of {DBMS:} a 10-year
  prediction for Enterprise-Grade {ML}}. In \bibinfo{booktitle}{\emph{{CIDR}}}.
\newblock


\bibitem[\protect\citeauthoryear{Bergstra and Bengio}{Bergstra and
  Bengio}{2012}]%
        {DBLP:journals/jmlr/BergstraB12}
\bibfield{author}{\bibinfo{person}{James Bergstra} {and}
  \bibinfo{person}{Yoshua Bengio}.} \bibinfo{year}{2012}\natexlab{}.
\newblock \showarticletitle{Random Search for Hyper-Parameter Optimization}.
\newblock \bibinfo{journal}{\emph{J. Mach. Learn. Res.}}  \bibinfo{volume}{13}
  (\bibinfo{year}{2012}), \bibinfo{pages}{281--305}.
\newblock


\bibitem[\protect\citeauthoryear{Chen and Guestrin}{Chen and Guestrin}{2016}]%
        {DBLP:conf/kdd/ChenG16}
\bibfield{author}{\bibinfo{person}{Tianqi Chen} {and} \bibinfo{person}{Carlos
  Guestrin}.} \bibinfo{year}{2016}\natexlab{}.
\newblock \showarticletitle{XGBoost: {A} Scalable Tree Boosting System}. In
  \bibinfo{booktitle}{\emph{Proceedings of the 22nd {ACM} {SIGKDD}
  International Conference on Knowledge Discovery and Data Mining}}.
\newblock


\bibitem[\protect\citeauthoryear{Das, Ivkin, Bansal, Rouesnel, Gautier, Karnin,
  Dirac, Ramakrishnan, Perunicic, Shcherbatyi, Wu, Zolic, Shen, Ahmed,
  Winkelmolen, Miladinovic, Archambeau, Tang, Dutt, Grao, and Venkateswar}{Das
  et~al\mbox{.}}{2020}]%
        {DBLP:conf/sigmod/DasIBRGKDRPSWZS20}
\bibfield{author}{\bibinfo{person}{Piali Das}, \bibinfo{person}{Nikita Ivkin},
  \bibinfo{person}{Tanya Bansal}, \bibinfo{person}{Laurence Rouesnel},
  \bibinfo{person}{Philip Gautier}, \bibinfo{person}{Zohar~S. Karnin},
  \bibinfo{person}{Leo Dirac}, \bibinfo{person}{Lakshmi Ramakrishnan},
  \bibinfo{person}{Andre Perunicic}, \bibinfo{person}{Iaroslav Shcherbatyi},
  \bibinfo{person}{Wilton Wu}, \bibinfo{person}{Aida Zolic},
  \bibinfo{person}{Huibin Shen}, \bibinfo{person}{Amr Ahmed},
  \bibinfo{person}{Fela Winkelmolen}, \bibinfo{person}{Miroslav Miladinovic},
  \bibinfo{person}{C{\'{e}}dric Archambeau}, \bibinfo{person}{Alex Tang},
  \bibinfo{person}{Bhaskar Dutt}, \bibinfo{person}{Patricia Grao}, {and}
  \bibinfo{person}{Kumar Venkateswar}.} \bibinfo{year}{2020}\natexlab{}.
\newblock \showarticletitle{Amazon SageMaker Autopilot: a white box AutoML
  solution at scale}. In \bibinfo{booktitle}{\emph{DEEM@SIGMOD}}.
\newblock


\bibitem[\protect\citeauthoryear{Erickson, Mueller, Shirkov, Zhang, Larroy, Li,
  and Smola}{Erickson et~al\mbox{.}}{2020}]%
        {agtabular}
\bibfield{author}{\bibinfo{person}{Nick Erickson}, \bibinfo{person}{Jonas
  Mueller}, \bibinfo{person}{Alexander Shirkov}, \bibinfo{person}{Hang Zhang},
  \bibinfo{person}{Pedro Larroy}, \bibinfo{person}{Mu Li}, {and}
  \bibinfo{person}{Alexander Smola}.} \bibinfo{year}{2020}\natexlab{}.
\newblock \showarticletitle{AutoGluon-Tabular: Robust and Accurate AutoML for
  Structured Data}.
\newblock \bibinfo{journal}{\emph{arXiv:2003.06505}} (\bibinfo{year}{2020}).
\newblock


\bibitem[\protect\citeauthoryear{Feurer, Eggensperger, Falkner, Lindauer, and
  Hutter}{Feurer et~al\mbox{.}}{2020}]%
        {DBLP:journals/corr/abs-2007-04074}
\bibfield{author}{\bibinfo{person}{Matthias Feurer}, \bibinfo{person}{Katharina
  Eggensperger}, \bibinfo{person}{Stefan Falkner}, \bibinfo{person}{Marius
  Lindauer}, {and} \bibinfo{person}{Frank Hutter}.}
  \bibinfo{year}{2020}\natexlab{}.
\newblock \showarticletitle{Auto-Sklearn 2.0: The Next Generation}.
\newblock  (\bibinfo{year}{2020}).
\newblock
\showeprint[arXiv]{2007.04074}


\bibitem[\protect\citeauthoryear{Feurer, Klein, Eggensperger, Springenberg,
  Blum, and Hutter}{Feurer et~al\mbox{.}}{2015}]%
        {DBLP:conf/nips/FeurerKESBH15}
\bibfield{author}{\bibinfo{person}{Matthias Feurer}, \bibinfo{person}{Aaron
  Klein}, \bibinfo{person}{Katharina Eggensperger},
  \bibinfo{person}{Jost~Tobias Springenberg}, \bibinfo{person}{Manuel Blum},
  {and} \bibinfo{person}{Frank Hutter}.} \bibinfo{year}{2015}\natexlab{}.
\newblock \showarticletitle{Efficient and Robust Automated Machine Learning}.
  In \bibinfo{booktitle}{\emph{{NIPS}}}.
\newblock


\bibitem[\protect\citeauthoryear{Gijsbers, LeDell, Poirier, Thomas, Bischl, and
  Vanschoren}{Gijsbers et~al\mbox{.}}{2019}]%
        {amlb2019}
\bibfield{author}{\bibinfo{person}{P. Gijsbers}, \bibinfo{person}{E. LeDell},
  \bibinfo{person}{S. Poirier}, \bibinfo{person}{J. Thomas},
  \bibinfo{person}{B. Bischl}, {and} \bibinfo{person}{J. Vanschoren}.}
  \bibinfo{year}{2019}\natexlab{}.
\newblock \showarticletitle{An Open Source AutoML Benchmark}.
\newblock \bibinfo{journal}{\emph{arXiv preprint arXiv:1907.00909 [cs.LG]}}
  (\bibinfo{year}{2019}).
\newblock
\newblock
\shownote{AutoML Workshop at ICML 2019.}


\bibitem[\protect\citeauthoryear{Gijsbers, Pfisterer, van Rijn, Bischl, and
  Vanschoren}{Gijsbers et~al\mbox{.}}{2021}]%
        {DBLP:conf/gecco/GijsbersPRBV21}
\bibfield{author}{\bibinfo{person}{Pieter Gijsbers}, \bibinfo{person}{Florian
  Pfisterer}, \bibinfo{person}{Jan~N. van Rijn}, \bibinfo{person}{Bernd
  Bischl}, {and} \bibinfo{person}{Joaquin Vanschoren}.}
  \bibinfo{year}{2021}\natexlab{}.
\newblock \showarticletitle{Meta-learning for symbolic hyperparameter
  defaults}. In \bibinfo{booktitle}{\emph{{GECCO} Companion}}.
\newblock


\bibitem[\protect\citeauthoryear{Guyon, Bennett, Cawley, Escalante, Escalera,
  Ho, Maci{\`{a}}, Ray, Saeed, Statnikov, and Viegas}{Guyon
  et~al\mbox{.}}{2015}]%
        {DBLP:conf/ijcnn/GuyonBCEEHMRSSV15}
\bibfield{author}{\bibinfo{person}{Isabelle Guyon}, \bibinfo{person}{Kristin~P.
  Bennett}, \bibinfo{person}{Gavin~C. Cawley}, \bibinfo{person}{Hugo~Jair
  Escalante}, \bibinfo{person}{Sergio Escalera}, \bibinfo{person}{Tin~Kam Ho},
  \bibinfo{person}{N{\'{u}}ria Maci{\`{a}}}, \bibinfo{person}{Bisakha Ray},
  \bibinfo{person}{Mehreen Saeed}, \bibinfo{person}{Alexander~R. Statnikov},
  {and} \bibinfo{person}{Evelyne Viegas}.} \bibinfo{year}{2015}\natexlab{}.
\newblock \showarticletitle{Design of the 2015 ChaLearn AutoML challenge}. In
  \bibinfo{booktitle}{\emph{{IJCNN}}}. \bibinfo{publisher}{{IEEE}},
  \bibinfo{pages}{1--8}.
\newblock


\bibitem[\protect\citeauthoryear{Hand and Till}{Hand and Till}{2001}]%
        {DBLP:journals/ml/HandT01}
\bibfield{author}{\bibinfo{person}{David~J. Hand} {and}
  \bibinfo{person}{Robert~J. Till}.} \bibinfo{year}{2001}\natexlab{}.
\newblock \showarticletitle{A Simple Generalisation of the Area Under the {ROC}
  Curve for Multiple Class Classification Problems}.
\newblock \bibinfo{journal}{\emph{Mach. Learn.}} \bibinfo{volume}{45},
  \bibinfo{number}{2} (\bibinfo{year}{2001}), \bibinfo{pages}{171--186}.
\newblock


\bibitem[\protect\citeauthoryear{Hutter, Hoos, and Leyton-Brown}{Hutter
  et~al\mbox{.}}{2011}]%
        {hutter2011}
\bibfield{author}{\bibinfo{person}{Frank Hutter}, \bibinfo{person}{Holger~H.
  Hoos}, {and} \bibinfo{person}{Kevin Leyton-Brown}.}
  \bibinfo{year}{2011}\natexlab{}.
\newblock \showarticletitle{Sequential Model-Based Optimization for General
  Algorithm Configuration}. In \bibinfo{booktitle}{\emph{Learning and
  Intelligent Optimization}}.
\newblock


\bibitem[\protect\citeauthoryear{Ke, Meng, Finley, Wang, Chen, Ma, Ye, and
  Liu}{Ke et~al\mbox{.}}{2017}]%
        {DBLP:conf/nips/KeMFWCMYL17}
\bibfield{author}{\bibinfo{person}{Guolin Ke}, \bibinfo{person}{Qi Meng},
  \bibinfo{person}{Thomas Finley}, \bibinfo{person}{Taifeng Wang},
  \bibinfo{person}{Wei Chen}, \bibinfo{person}{Weidong Ma},
  \bibinfo{person}{Qiwei Ye}, {and} \bibinfo{person}{Tie{-}Yan Liu}.}
  \bibinfo{year}{2017}\natexlab{}.
\newblock \showarticletitle{LightGBM: {A} Highly Efficient Gradient Boosting
  Decision Tree}. In \bibinfo{booktitle}{\emph{Advances in Neural Information
  Processing Systems 30}}.
\newblock


\bibitem[\protect\citeauthoryear{Lavesson and Davidsson}{Lavesson and
  Davidsson}{2006}]%
        {DBLP:conf/aaai/LavessonD06}
\bibfield{author}{\bibinfo{person}{Niklas Lavesson} {and} \bibinfo{person}{Paul
  Davidsson}.} \bibinfo{year}{2006}\natexlab{}.
\newblock \showarticletitle{Quantifying the Impact of Learning Algorithm
  Parameter Tuning}. In \bibinfo{booktitle}{\emph{{AAAI}}}.
\newblock


\bibitem[\protect\citeauthoryear{LeDell and Poirier}{LeDell and
  Poirier}{2020}]%
        {ledell2020h2o}
\bibfield{author}{\bibinfo{person}{Erin LeDell} {and}
  \bibinfo{person}{Sebastien Poirier}.} \bibinfo{year}{2020}\natexlab{}.
\newblock \showarticletitle{H2o automl: Scalable automatic machine learning}.
  In \bibinfo{booktitle}{\emph{Proceedings of the AutoML Workshop at ICML}}.
\newblock


\bibitem[\protect\citeauthoryear{Olson, Urbanowicz, Andrews, Lavender, Kidd,
  and Moore}{Olson et~al\mbox{.}}{2016}]%
        {Olson2016TPOT}
\bibfield{author}{\bibinfo{person}{Randal~S. Olson}, \bibinfo{person}{Ryan~J.
  Urbanowicz}, \bibinfo{person}{Peter~C. Andrews}, \bibinfo{person}{Nicole~A.
  Lavender}, \bibinfo{person}{La~Creis Kidd}, {and} \bibinfo{person}{Jason~H.
  Moore}.} \bibinfo{year}{2016}\natexlab{}.
\newblock \showarticletitle{Automating Biomedical Data Science Through
  Tree-Based Pipeline Optimization}. In \bibinfo{booktitle}{\emph{Applications
  of Evolutionary Computation}}, \bibfield{editor}{\bibinfo{person}{Giovanni
  Squillero} {and} \bibinfo{person}{Paolo Burelli}} (Eds.).
  \bibinfo{publisher}{Springer International Publishing},
  \bibinfo{pages}{123--137}.
\newblock


\bibitem[\protect\citeauthoryear{Pedregosa, Varoquaux, Gramfort, Michel,
  Thirion, Grisel, Blondel, Prettenhofer, Weiss, Dubourg, Vanderplas, Passos,
  Cournapeau, Brucher, Perrot, and Duchesnay}{Pedregosa et~al\mbox{.}}{2011}]%
        {scikit-learn}
\bibfield{author}{\bibinfo{person}{F. Pedregosa}, \bibinfo{person}{G.
  Varoquaux}, \bibinfo{person}{A. Gramfort}, \bibinfo{person}{V. Michel},
  \bibinfo{person}{B. Thirion}, \bibinfo{person}{O. Grisel},
  \bibinfo{person}{M. Blondel}, \bibinfo{person}{P. Prettenhofer},
  \bibinfo{person}{R. Weiss}, \bibinfo{person}{V. Dubourg}, \bibinfo{person}{J.
  Vanderplas}, \bibinfo{person}{A. Passos}, \bibinfo{person}{D. Cournapeau},
  \bibinfo{person}{M. Brucher}, \bibinfo{person}{M. Perrot}, {and}
  \bibinfo{person}{E. Duchesnay}.} \bibinfo{year}{2011}\natexlab{}.
\newblock \showarticletitle{Scikit-learn: Machine Learning in {P}ython}.
\newblock \bibinfo{journal}{\emph{Journal of Machine Learning Research}}
  \bibinfo{volume}{12} (\bibinfo{year}{2011}), \bibinfo{pages}{2825--2830}.
\newblock


\bibitem[\protect\citeauthoryear{Prokhorenkova, Gusev, Vorobev, Dorogush, and
  Gulin}{Prokhorenkova et~al\mbox{.}}{2018}]%
        {DBLP:conf/nips/ProkhorenkovaGV18}
\bibfield{author}{\bibinfo{person}{Liudmila~Ostroumova Prokhorenkova},
  \bibinfo{person}{Gleb Gusev}, \bibinfo{person}{Aleksandr Vorobev},
  \bibinfo{person}{Anna~Veronika Dorogush}, {and} \bibinfo{person}{Andrey
  Gulin}.} \bibinfo{year}{2018}\natexlab{}.
\newblock \showarticletitle{CatBoost: unbiased boosting with categorical
  features}. In \bibinfo{booktitle}{\emph{Advances in Neural Information
  Processing Systems 31}}.
\newblock


\bibitem[\protect\citeauthoryear{Shang, Zgraggen, Buratti, Kossmann, Eichmann,
  Chung, Binnig, Upfal, and Kraska}{Shang et~al\mbox{.}}{2019}]%
        {DBLP:conf/sigmod/ShangZBKECBUK19}
\bibfield{author}{\bibinfo{person}{Zeyuan Shang}, \bibinfo{person}{Emanuel
  Zgraggen}, \bibinfo{person}{Benedetto Buratti}, \bibinfo{person}{Ferdinand
  Kossmann}, \bibinfo{person}{Philipp Eichmann}, \bibinfo{person}{Yeounoh
  Chung}, \bibinfo{person}{Carsten Binnig}, \bibinfo{person}{Eli Upfal}, {and}
  \bibinfo{person}{Tim Kraska}.} \bibinfo{year}{2019}\natexlab{}.
\newblock \showarticletitle{Democratizing Data Science through Interactive
  Curation of {ML} Pipelines}. In \bibinfo{booktitle}{\emph{{SIGMOD}
  Conference}}.
\newblock


\bibitem[\protect\citeauthoryear{Singh, Kates, Mentch, Kharkar, Udell, and
  Drori}{Singh et~al\mbox{.}}{2021}]%
        {singh2021privileged}
\bibfield{author}{\bibinfo{person}{Nikhil Singh}, \bibinfo{person}{Brandon
  Kates}, \bibinfo{person}{Jeff Mentch}, \bibinfo{person}{Anant Kharkar},
  \bibinfo{person}{Madeleine Udell}, {and} \bibinfo{person}{Iddo Drori}.}
  \bibinfo{year}{2021}\natexlab{}.
\newblock \showarticletitle{Privileged Zero-Shot AutoML}.
\newblock \bibinfo{journal}{\emph{arXiv preprint arXiv:2106.13743}}
  (\bibinfo{year}{2021}).
\newblock


\bibitem[\protect\citeauthoryear{Wang, Wu, Weimer, and Zhu}{Wang
  et~al\mbox{.}}{2021}]%
        {wang2021flaml}
\bibfield{author}{\bibinfo{person}{Chi Wang}, \bibinfo{person}{Qingyun Wu},
  \bibinfo{person}{Markus Weimer}, {and} \bibinfo{person}{Erkang Zhu}.}
  \bibinfo{year}{2021}\natexlab{}.
\newblock \showarticletitle{FLAML: A Fast and Lightweight AutoML Library}. In
  \bibinfo{booktitle}{\emph{MLSys}}.
\newblock


\bibitem[\protect\citeauthoryear{Wolpert and Macready}{Wolpert and
  Macready}{[n.d.]}]%
        {DBLP:journals/tec/DolpertM97}
\bibfield{author}{\bibinfo{person}{David~H. Wolpert} {and}
  \bibinfo{person}{William~G. Macready}.} \bibinfo{year}{[n.d.]}\natexlab{}.
\newblock \showarticletitle{No free lunch theorems for optimization}.
\newblock \bibinfo{journal}{\emph{{IEEE} Trans. Evol. Comput.}}
  \bibinfo{number}{1} (\bibinfo{year}{[n.\,d.]}), \bibinfo{pages}{67--82}.
\newblock


\bibitem[\protect\citeauthoryear{Wu, Wang, Langford, Mineiro, and Rossi}{Wu
  et~al\mbox{.}}{2021}]%
        {pmlr-v139-wu21d}
\bibfield{author}{\bibinfo{person}{Qingyun Wu}, \bibinfo{person}{Chi Wang},
  \bibinfo{person}{John Langford}, \bibinfo{person}{Paul Mineiro}, {and}
  \bibinfo{person}{Marco Rossi}.} \bibinfo{year}{2021}\natexlab{}.
\newblock \showarticletitle{ChaCha for Online AutoML}. In
  \bibinfo{booktitle}{\emph{Proceedings of the 38th International Conference on
  Machine Learning}}.
\newblock


\bibitem[\protect\citeauthoryear{Yakovlev, Moghadam, Moharrer, Cai, Chavoshi,
  Varadarajan, Agrawal, Karnagel, Idicula, Jinturkar, and Agarwal}{Yakovlev
  et~al\mbox{.}}{2020}]%
        {DBLP:journals/pvldb/YakovlevMMCCVAK20}
\bibfield{author}{\bibinfo{person}{Anatoly Yakovlev},
  \bibinfo{person}{Hesam~Fathi Moghadam}, \bibinfo{person}{Ali Moharrer},
  \bibinfo{person}{Jingxiao Cai}, \bibinfo{person}{Nikan Chavoshi},
  \bibinfo{person}{Venkatanathan Varadarajan}, \bibinfo{person}{Sandeep~R.
  Agrawal}, \bibinfo{person}{Tomas Karnagel}, \bibinfo{person}{Sam Idicula},
  \bibinfo{person}{Sanjay Jinturkar}, {and} \bibinfo{person}{Nipun Agarwal}.}
  \bibinfo{year}{2020}\natexlab{}.
\newblock \showarticletitle{Oracle AutoML: {A} Fast and Predictive AutoML
  Pipeline}.
\newblock \bibinfo{journal}{\emph{Proc. {VLDB} Endow.}} \bibinfo{volume}{13},
  \bibinfo{number}{12} (\bibinfo{year}{2020}), \bibinfo{pages}{3166--3180}.
\newblock


\end{thebibliography}

\onecolumn
\section*{Appendix}

\begin{longtable}{  l| l |r| l | r | r} 
\caption{Metafeatures.} \label{tab:data}
\centering
\small
\cr
name & task id & \# instance  & \# feature & \# class & \% numeric feature\\
\hline
Australian & 146818 & 621 & 14 & 2 & 0.428571429 \\
blood-transfusion & 10101 & 674 & 4 & 2 & 1 \\
car & 146821 & 1556 & 6 & 4 & 0 \\
christine & 168908 & 4877 & 1611 & 2 & 0.99255121 \\
cnae-9 & 9981 & 972 & 807 & 9 & 1 \\
credit-g & 31 & 900 & 20 & 2 & 0.35 \\
dilbert & 168909 & 9000 & 2000 & 5 & 1 \\
fabert & 168910 & 7414 & 795 & 7 & 1 \\
jasmine & 168911 & 2686 & 144 & 2 & 0.055555556 \\
kc1 & 3917 & 1899 & 21 & 2 & 1 \\
kr-vs-kp & 3 & 2877 & 36 & 2 & 0 \\
mfeat-factors & 12 & 1800 & 216 & 10 & 1 \\
phoneme & 9952 & 4864 & 5 & 2 & 1 \\
segment & 146822 & 2079 & 16 & 7 & 1 \\
sylvine & 168912 & 4612 & 20 & 2 & 1 \\
vehicle & 53 & 762 & 18 & 4 & 1 \\
adult & 7592 & 43958 & 14 & 2 & 0.428571429 \\
Amazon\_employee\_access & 34539 & 29493 & 9 & 2 & 0 \\
APSFailure & 168868 & 68400 & 169 & 2 & 1 \\
bank-marketing & 14965 & 40690 & 16 & 2 & 0.4375 \\
connect-4 & 146195 & 60802 & 42 & 3 & 0 \\
Fashion-MNIST & 146825 & 63000 & 784 & 10 & 1 \\
guillermo & 168337 & 18000 & 4281 & 2 & 1 \\
Helena & 168329 & 58677 & 27 & 100 & 1 \\
higgs & 146606 & 88245 & 28 & 2 & 1 \\
Jannis & 168330 & 75360 & 54 & 4 & 1 \\
jungle\_chess\_2pcs\_raw\_endgame\_complete & 167119 & 40338 & 6 & 3 & 1 \\
KDDCup09\_appetency & 3945 & 45000 & 207 & 2 & 0.835748792 \\
MiniBooNE & 168335 & 117058 & 50 & 2 & 1 \\
nomao & 9977 & 31019 & 118 & 2 & 0.754237288 \\
numerai28\_6 & 167120 & 86688 & 21 & 2 & 1 \\
riccardo & 168338 & 18000 & 4283 & 2 & 1 \\
Robert & 168332 & 9000 & 7200 & 10 & 1 \\
Shuttle & 146212 & 52200 & 9 & 7 & 1 \\
Volkert & 168331 & 52479 & 147 & 10 & 1 \\
Airlines & 189354 & 485445 & 7 & 2 & 0.428571429 \\
Albert & 189356 & 382716 & 78 & 2 & 0.333333333 \\
Covertype & 7593 & 522911 & 54 & 7 & 0.185185185 \\
Dionis & 189355 & 374570 & 54 & 355 & 1 \\
poker & 10102 & 922509 & 10 & 0 & 1 \\
pol & 2292 & 13500 & 26 & 0 & 1 \\
2dplanes & 2306 & 36692 & 10 & 0 & 1 \\
bng\_breastTumor & 7324 & 104976 & 9 & 0 & 0.111111111 \\
bng\_echomonths & 7323 & 15747 & 9 & 0 & 0.666666667 \\
bng\_lowbwt & 7320 & 27994 & 9 & 0 & 0.222222222 \\
bng\_pbc & 7318 & 900000 & 18 & 0 & 0.555555556 \\
bng\_pharynx & 7322 & 900000 & 10 & 0 & 0.1 \\
bng\_pwLinear & 7325 & 159433 & 10 & 0 & 1 \\
fried & 4885 & 36692 & 10 & 0 & 1 \\
house\_16H & 4893 & 20506 & 16 & 0 & 1 \\
house\_8L & 2309 & 20506 & 8 & 0 & 1 \\
houses & 5165 & 18576 & 8 & 0 & 1 \\
mv & 4774 & 36692 & 10 & 0 & 0.7 \\
cpu\_act & 4892 & 7373 & 21 & 0 & 1 \\
bng\_autoPrice & 7321 & 900000 & 15 & 0 & 1 \\
elevators & 2307 & 14940 & 18 & 0 & 1 \\
house\_sales & 359949 & 19452 & 21 & 0 & 0.952380952 \\
brazil\_houses & 359938 & 9623 & 12 & 0 & 0.666666667 \\
bng\_satellite & 7326 & 900000 & 36 & 0 & 1 \\
airlines\_depdelay & 359926 & 900000 & 9 & 0 & 0.666666667 \\
allstate\_claims & 233212 & 169487 & 130 & 0 & 0.107692308 \\
black\_friday & 168891 & 150139 & 9 & 0 & 0.555555556 \\
buzz\_twitter & 233213 & 524925 & 77 & 0 & 1 \\
ailerons & 4769 & 12375 & 40 & 0 & 1 \\
comet & 14949 & 6857460 & 4 & 0 & 1 \\
rainfall\_bangladesh & 168889 & 15080 & 3 & 0 & 0.333333333 \\
bng\_libras\_move & 7327 & 900000 & 90 & 0 & 1 \\
dataset\_sales & 190418 & 9665 & 14 & 0 & 1 \\
diamonds & 233211 & 48546 & 9 & 0 & 0.666666667 \\
Yolanda & 317614 & 360000 & 100 & 0 & 1 \\
nyc\_taxi & 359943 & 523652 & 18 & 0 & 0.5 \\
bike\_sharing\_demand & 317615 & 15642 & 12 & 0 & 0.666666667 \\
sulfur & 360966 & 9073 & 6 & 0 & 1 \\
nasa\_phm & 360879 & 41327 & 17 & 0 & 1 \\
particulate-matter & 360968 & 354870 & 9 & 0 & 0.333333333 \\
miami\_houses & 360969 & 12539 & 16 & 0 & 1 \\
\end{longtable}

\vspace{-0.75em}
\subsection{Datasets}

We selected datasets available through the Open\textsc{ml} online repository. Binary and multiclass datasets were sourced from the AutoMLBenchmark framework~\cite{amlb2019}, which are specifically selected to be varied and challenging in the auto\textsc{ml} setting. We augmented these with regression datasets used as a benchmark in previous auto\textsc{ml} work~\cite{pmlr-v139-wu21d}. These regression tasks were selected from Open\textsc{ml} using the criteria that they contained at least ten thousand rows, had no missing values and were active and downloadable. Additionally, we compiled 10 extra regression tasks found on OpenML, which did not necessarily meet the these criteria. We used these additional tasks in the application section, Section~\ref{sec:application}, but not during the evaluation against other approaches in Section~\ref{section:results}. Finally, five classification tasks failed to run in the testing framework---mostly likely due to an invalid specification in the online repository---and so were excluded from our report. These are: kc1, MiniBooNE, kr-vs-kp, connect-4 and Dionis.

Cross-validation datasets: 
christine, 
Airlines, 
bank-marketing, 
Australian, 
kr-vs-kp, 
blood-transfusion, 
riccardo, 
Albert, 
APSFailure, 
higgs, 
nomao, 
adult, 
MiniBooNE, 
numerai28\_6, 
kc1, 
credit-g, 
connect-4, 
Shuttle, 
jungle\_chess\_2pcs\_raw\_endgame\_complete, 
Fashion-MNIST, 
mfeat-factors, 
cnae-9, 
Helena, 
car, 
fabert, 
vehicle, 
Covertype, 
Dionis, 
Robert, 
dilbert, 
pol,
2dplanes,
bng\_echomonths,
bng\_lowbwt,
bng\_pbc,
bng\_pharynx,
fried,
houses,
mv,
cpu\_act,
bng\_autoPrice,
elevators,
house\_sales,
brazil\_houses,
bng\_satellite,
airlines\_depdelay,
allstate\_claims,
black\_friday,
buzz\_twitter,
ailerons,
comet,
rainfall\_bangladesh.

Holdout datasets:
jasmine,
guillermo,
KDDCup09\_appetency,
Amazon\_employee\_access,
phoneme,
sylvine,
segment,
Volkert,
Jannis,
poker,
bng\_breastTumor,
bng\_pwLinear,
house\_16H,
house\_8L,
bng\_autoHorse,
online\_news,
new\_fuel\_car,
mnist\_rotation.

The corresponding Open\textsc{ml} task ids and metafeatures are presented in Table~\ref{tab:data}.

\subsection{Experiment details}

The evaluation of each method is based on the open source benchmark code at \url{https://github.com/openml/automlbenchmark}.

For the offline mining step, we utilized a server with 128 GB of main memory and a 16 core processor clocked at 2.8 GHz, providing 32 simultaneous multi-threading threads.
We set an offline evaluation budget of 100 minutes (10 minutes over 10 folds) per task, and an online budget of one minute for model selection followed by one minute for model training. We set the target regret to $\varepsilon = 0.01$. We did not tune the target regret parameter, though one could tune it via cross validation. For the online step, we utilized a single thread on the same processor with a 4 GB memory pool.

As the performance of any auto\textsc{ml} method is heavily influenced by the choice of search space, to make fair comparisons between frameworks we re-implemented AutoSklearn 1.0 and 2.0 to use the same search space as \textsc{Flaml}. Note that this change is in favor of the baseline because \textsc{Flaml}'s search space contains more powerful learners like lightgbm and xgboost than sklearn. Our extension of AutoSklearn 1.0 which enables knowledge transfer across tasks recommends the best configuration on the nearest neighbor of an input task among all the candidate configurations.

The number of models trained in the k-shot \textsc{Askl} 2.0 in the 1-minute time budget ranges from 3 to 7, depending on the training time of each model.

\end{document}